\documentclass[10pt,twocolumn,letterpaper]{article}

\usepackage{iccv}
\usepackage{times}
\usepackage{epsfig}
\usepackage{graphicx}
\usepackage{amsmath}
\usepackage{amssymb}

\usepackage[pagebackref=true,breaklinks=true,letterpaper=true,colorlinks,bookmarks=false]{hyperref}

\iccvfinalcopy 

\def\iccvPaperID{33} 

\ificcvfinal\fi

\begin{document}

\title{On-device Real-time Custom Hand Gesture Recognition}

\author{
Esha Uboweja\quad David Tian\quad Qifei Wang\quad Yi-Chun Kuo\quad Joe Zou\quad Lu Wang\\
George Sung\quad Matthias Grundmann\\
Google LLC\\
1600 Amphitheatre Pkway, Mountain View, CA 94043, USA\\
{\tt\small \{eshauboweja, dctian, qfwang, yichunkuo, zouj, luwa, gsung, grundman\}@google.com}\\
}

\maketitle
\ificcvfinal\thispagestyle{empty}\fi

\begin{abstract}
Most existing hand gesture recognition (HGR) systems are limited to a predefined set of gestures.
However, users and developers often want to recognize new, unseen gestures.
This is challenging due to the vast diversity of all plausible hand shapes, \eg it is impossible for developers to include all hand gestures in a predefined list.

In this paper, we present a user-friendly framework that lets users easily customize and deploy their own gesture recognition pipeline.
Our framework provides a pre-trained single-hand embedding model that can be fine-tuned for custom gesture recognition.
Users can perform gestures in front of a webcam to collect a small amount of images per gesture.
We also offer a low-code solution to train and deploy the custom gesture recognition model.
This makes it easy for users with limited ML expertise to use our framework.
We further provide a no-code web front-end for users without any ML expertise.
This makes it even easier to build and test the end-to-end pipeline.
The resulting custom HGR is then ready to be run on-device for real-time scenarios.
This can be done by calling a simple function in our open-sourced model inference API,
{\normalfont MediaPipe Tasks}.
This entire process only takes a few minutes.

\end{abstract}
\begin{figure}
    \centering
    \includegraphics[width=0.86\linewidth]{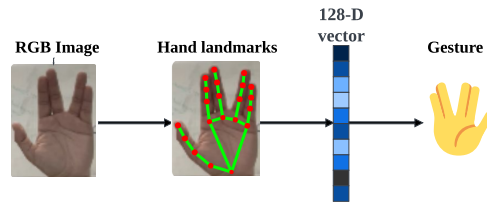}
    \caption{Our custom hand gesture recognition system enables any user without ML expertise to use a small number of images per gesture class for training and immediately use the model for real-time on-device inference.
    Here we show how our solution extracts the hand landmarks of each hand to compute a 128 dimensional embedding vector which is used for custom gesture classification.
    (The landmarks in this figure are best viewed digitally).
    }
    \label{fig:teaser}
\end{figure}
\section{Introduction}
Hand gesture recognition (HGR) plays a pivotal role in enabling natural and intuitive human-computer interactions, such as in augmented reality (AR), virtual reality (VR), video conferencing and remote control applications.
As these technologies evolve, the ability to accurately detect, interpret and respond to hand gestures is key to creating immersive user experiences without disruption.

We present an innovative approach to train accurate and robust HGR models with limited training data.
Our approach uses a pre-trained model that has been trained on a large dataset of videos of people fingerspelling words in sign language.
We then fine-tune the weights of this pre-trained model for custom gesture classification (see Figure~\ref{fig:teaser}).
This approach has two main benefits:
\begin{enumerate}
    \itemsep0em
    \item We are able to train an accurate model with a relatively small amount of training data, as few as $50$ images per gesture.
    \item The pre-trained model captures information of a wider range of hand shapes and movements, including transition states that are harder to capture with still images.
\end{enumerate}

Our HGR inference pipeline works as follows:
\begin{enumerate}
    \itemsep0em
    \item An RGB camera captures an image.
    \item The HGR extracts the $3$D skeletal key points (or landmarks) and the handedness (left, right) of each hand from the input image.
    \item The landmarks and handedness information are supplied to the newly trained custom gesture recognition model for inference.
\end{enumerate}

Our HGR runs real-time at 30+ FPS (frames per second) on mainstream mobile devices.


\section {Architecture}

We use the work presented in ``On-device Real-Time Hand Gesture Recognition’’~\cite{gesture2021} as the starting point for building a system for custom hand gesture recognition.
As shown in Figure~\ref{fig:teaser}, our solution uses a model that extracts hand landmarks and runs in real-time~\cite{mediapipe_hands_paper}.

\begin{figure}
    \centering
    \includegraphics[width=1\linewidth]{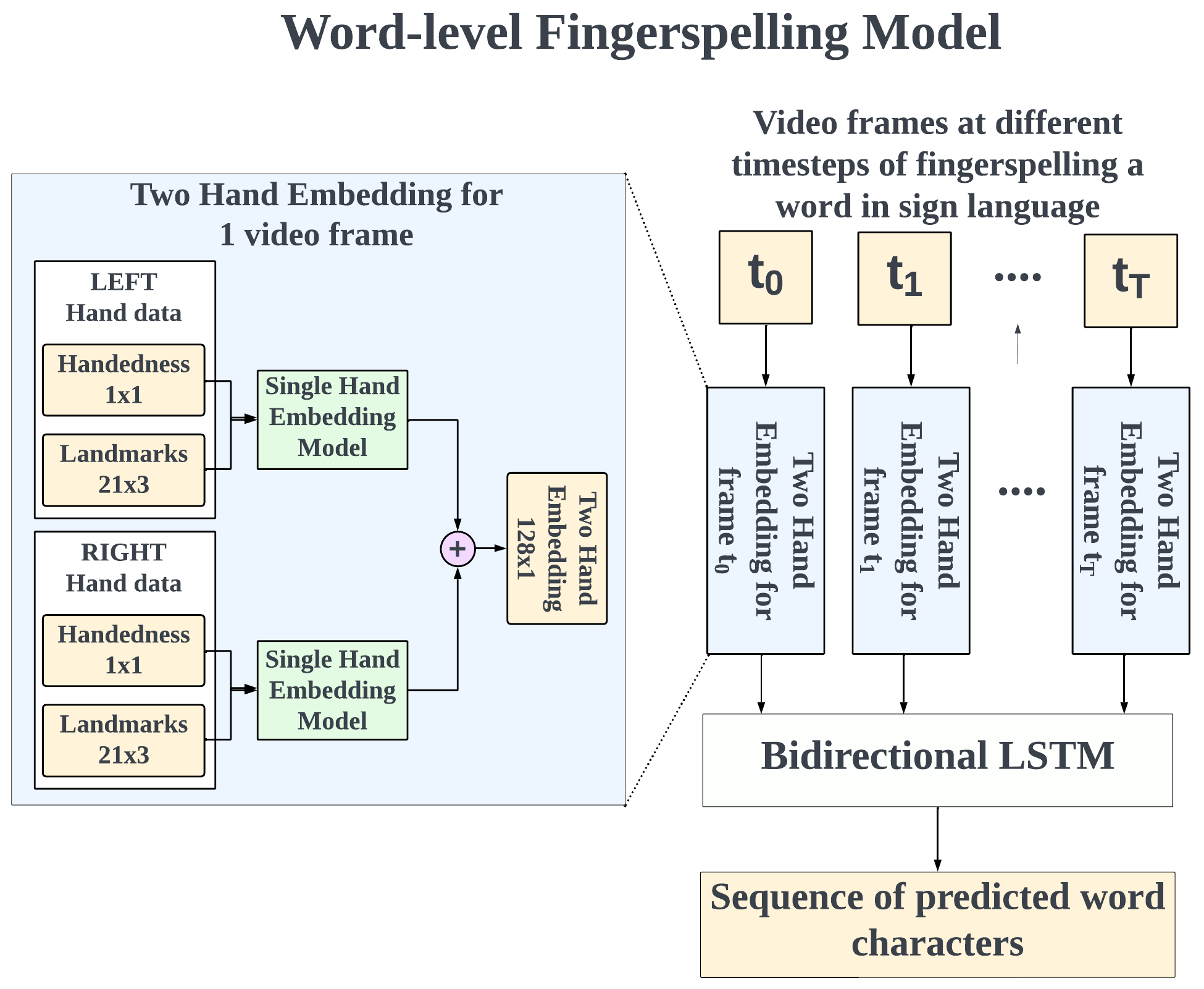}
    \caption{Model architecture used for training the \textit{word-level fingerspelling} model and the \textit{single-hand embedding} sub-model.}
    \label{fig:word_asl_training}
\end{figure}

To train our \textit{word-level fingerspelling} model, we use an in-house collected dataset of $79K$ videos of $21K$ unique fingerspelled words.
In each video, a subject fingerspells a word using either the left or the right hand.
During training we discard frames that don't contain any hands.
We use normalized hand landmarks after processing the input videos using the hand landmark model~\cite{mediapipe_hands_paper}.

As shown in Figure~\ref{fig:word_asl_training}, the \textit{word-level fingerspelling} model extracts embedding vectors from each hand's landmarks in each video frame.
Since each frame contains only one of \textit{left}, \textit{right} hands, the model extracts a \textit{single-hand embedding} for each hand and adds the two embedding vectors.
Since addition is commutative, the model is invariable to the order of the two embedding vectors.
The embedding vector of a single video frame contains structural piece-wise skeletal information.
All per-frame embedding vectors along with hand location information are sent to a lightweight bidirectional LSTM~\cite{lstm_paper}~\cite{bidirectional} to predict character level sequences of the fingerspelled word. 

Using a Connectionist Temporal Classification (CTC) loss~\cite{ctc_loss_paper} for training the \textit{word-level fingerspelling} model in Figure~\ref{fig:word_asl_training}, we are able to guide the \textit{single-hand embedding} sub-model to extract discriminative features that capture the subtle differences in a wide range of real-world hand configurations.
We are thus able to use weights of the \textit{single-hand embedding} model for training a custom gesture recognition model with minimal training data via transfer learning~\cite{transfer_learning_2023}.

\begin{figure}[hbt!]
\begin{center}
   \includegraphics[width=1\linewidth]{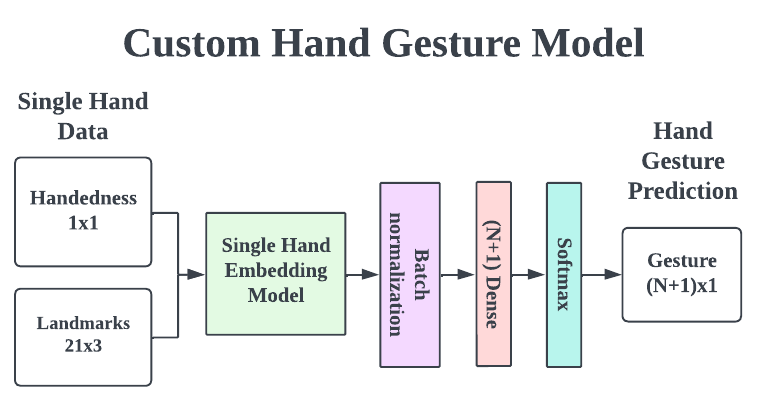}
\end{center}
   \caption{Custom hand gesture model.
   This model classifies the input hand data into one of $N+1$ classes,($N$ gestures and $1$ \textit{background} class).}
\label{fig:custom_gesture_model}
\end{figure}

We propose that the weights of the pre-trained \textit{single-hand embedding} model represent essential features that are useful for custom gesture recognition.
By fine-tuning the weights of the pre-trained embedding model and the custom hand gesture model head, we observe that our model can recognize gestures accurately.
This approach significantly reduces number of images required for training. 
Figure~\ref{fig:custom_gesture_model} shows the model architecture of the custom hand gesture recognition model with the \textit{single-hand embedding} model as its feature extractor.

\section {Results}

\begin{figure}
\begin{center}
   \includegraphics[width=1\linewidth]{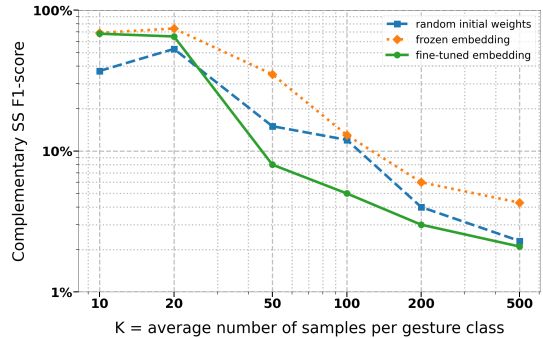}
\end{center}
   \caption{\textit{Complementary} $SS\ F_1 score$ for K-shot gesture classification.}
\label{fig:custom_gesture_performance}
\end{figure}

In Figure~\ref{fig:custom_gesture_performance}, we report the results of training a custom gesture recognition model by fine-tuning the weights of the \textit{single-hand embedding} model (shown in Figure~\ref{fig:custom_gesture_model}).

We used an in-house dataset of $8$ classes, with $7$ gesture classes and $1$ \textit{background} class. Samples that could not be labeled as any of the $7$ gesture classes were labeled as the \textit{background} class.
To explore how much data is required to train the custom gesture recognition model, we conducted trials with varying values of the average number of training samples per gesture, $K$.
We used the following values of $K: 10, 20, 50, 100, 200, 500$. For example, when $K = 20$, we train a model with $20$ positive and negative samples of each of the $N$ gesture classes, \ie the total number of samples used for training were $N \times 20$ ($140$ for $7$ gesture classes). The negative samples are labeled as the \textit{background} class.

During inference, an input hand shape can be labeled as one of the $8$ classes. We report the performance on the $7$ gesture classes. To account for the performance of the \textit{background} class in our results, we use \textit{specificity} and \textit{sensitivity}:
\begin{align*}
\mathrm{Specificity} = \frac{\mathrm{True\ Negatives}}{\mathrm{True\ Negatives} + \mathrm{False\ Positives}} \\
\mathrm{Sensitivity} = \frac{\mathrm{True\ Positives}}{\mathrm{True\ Positives} + \mathrm{False\ Negatives}}
\end{align*}
\textit{True negatives} account for samples that are correctly labeled as the \textit{background} class. Similarly, \textit{false positives} account for samples that belong to the \textit{background} class but are incorrectly labeled as one of the gesture classes.
To concisely represent our results, we combine \textit{sensitivity} and \textit{specificity} into one metric, namely the $\mathrm{SS\ F_1 score}$ which is the harmonic mean of these two metrics:
\begin{align*}
    \mathrm{SS\ F_1 score} = \frac{2 \times \mathrm{Sensitivity} \times \mathrm{Specificity}}{\mathrm{Sensitivity} + \mathrm{Specificity}}
\end{align*}
Most of our models achieve $\mathrm{SS\ F_1 score}$ values close to $1.0$. So we present results in Figure~\ref{fig:custom_gesture_performance} as
$$\mathrm{complementary\ SS\ F}_1\mathrm{score} =1 - \mathrm{SS\ F_1 score}$$
This allows us to measure the model's performance by focusing on misclassification errors.

To explore the effectiveness of the \textit{fine-tuned embedding} model for custom gesture classification, we conducted an ablation study on the model's weights. The models we trained for the study have the same architecture as the \textit{fine-tuned embedding} custom gesture model. We defined two experiments:
\begin{enumerate}
    \itemsep0em
    \item \textit{Random initial weights}: The initial weights of all layers are randomized, so the model trains on raw hand landmark data from scratch.
    \item \textit{Frozen embedding}: The weights of the \textit{single-hand embedding} model layers are frozen. Only the weights of the classification head are updated during training.
\end{enumerate}
We report the results of $K$-shot gesture classification for these models in Figure~\ref{fig:custom_gesture_performance}.
All models perform reasonably well when the value of $K$ is high, \ie $K = 500$.
For very small values of $K$, \ie $K = 10$ and $K = 20$, all models perform poorly.
Note that the model with random initial weights performs well for these values of $K$ but the \textit{complementary} $SS\ F_1score$ is still unacceptably much higher than $10\%$.

For values of $K = 50$ and above, we observe that the \textit{complementary} $SS\ F_1score$ is lower than $10\%$, steadily decreasing for higher values of $K$. The \textit{fine-tuned embedding} model outperforms the other two models at $K = 50$, $K = 100$ and $K = 200$. These results demonstrate the advantage of fine-tuning a pre-trained \textit{single-hand embedding} model instead of training a model with random initial weights to recognize hand gestures from raw hand landmarks.

\section {Hand Landmark Detection Improvements}

\begin{figure}
    \centering
    \includegraphics[width=1\linewidth]{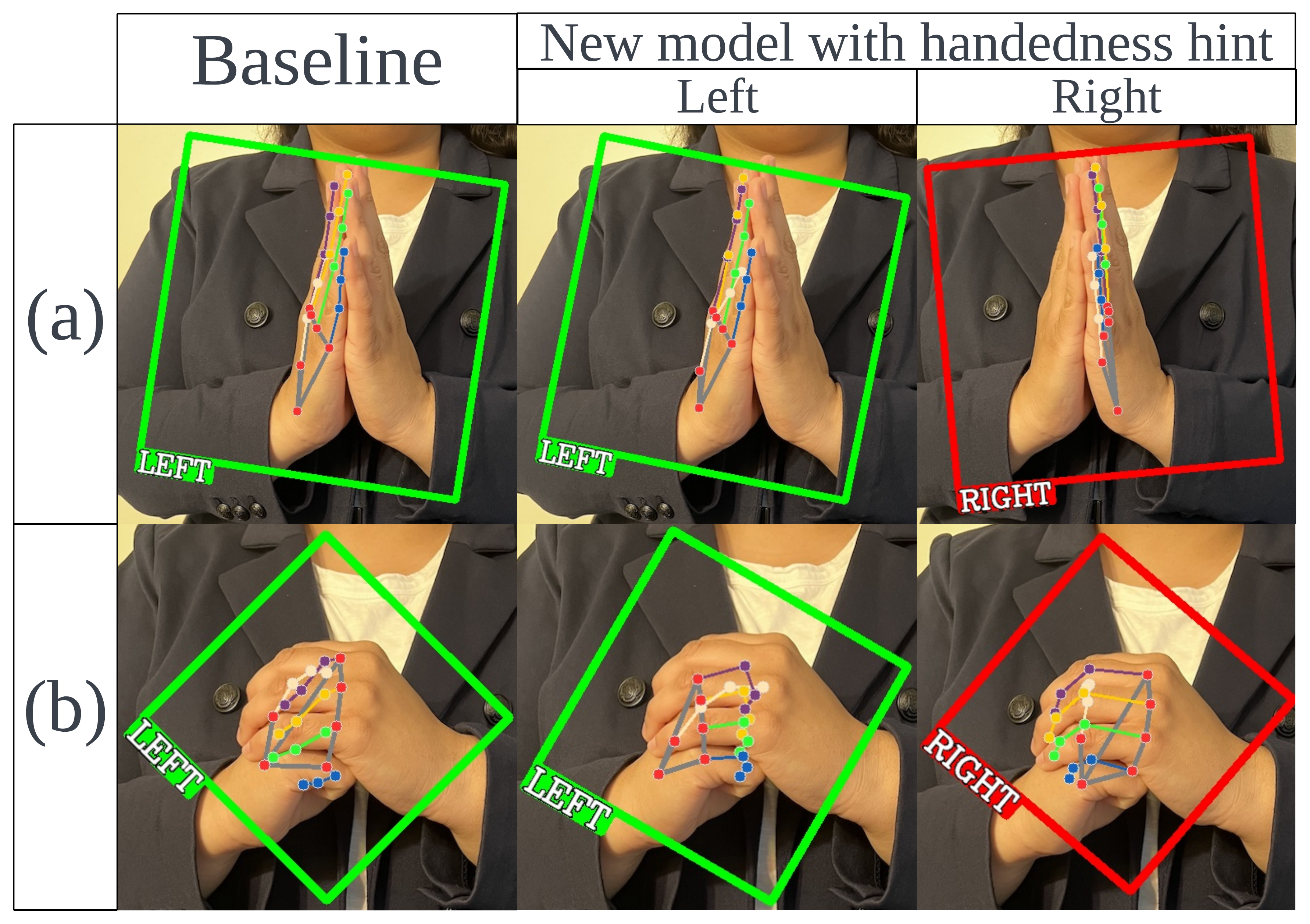}
    \caption{Results of using a hand landmark model with a handedness hint input compared to using the baseline hand landmark model without a handedness hint input.}
    \label{fig:handedness_hint}
\end{figure}

When two hands are very close to each other or occlude each other, the landmark model fails to accurately extract all hand landmarks for both hands. This failure cascades to the gesture recognition system that relies on accurate landmark detection to correctly infer the gesture depicted by a hand shape.
In Figure~\ref{fig:handedness_hint} for example, we can see that the baseline hand landmark model is unable to extract landmarks of the right hand in panels (a) and (b).

To improve landmark accuracy when two hands are near each other or are overlapping with each other, we experimented with providing a handedness hint to the hand landmark model during training and inference.
This guides the model to extract the landmarks of the hand with the same handedness as the input handedness hint.

In Figure~\ref{fig:handedness_hint}, we can see that the new model extracts both the left and right hand's landmarks respectively with the correct handedness hint.

Quantitatively, on an in-house dataset of $3,310$ images where hands are near or overlapping each other, the new model has a Mean Normalized Absolute Error (MNAE) of $13.09$ compared to the baseline model's MNAE of $13.89$. This improvement enables the custom hand gesture recognition pipeline to perform well when multiple hands are present in the input image or video.

\section {Implementation}

\begin{figure}[!htp]
    \begin{center}
        \includegraphics[width=1\linewidth]{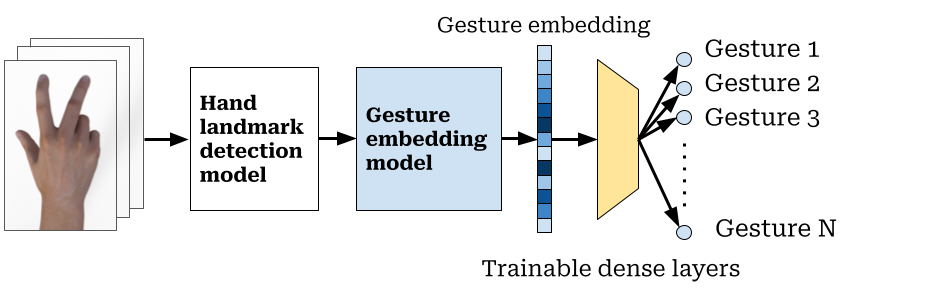}
    \end{center}
\caption{Training and Inference Pipelines.}
\label{fig:model_maker_pipeline}
\end{figure}

\subsection{Training Pipeline}

We developed a low-code training pipeline called \textit{MediaPipe Model Maker}~\cite{mediapipe_model_maker},
that enables users to effortlessly train new hand gesture recognition models.
In the pipeline, the custom gesture recognition model is defined as a set of dense layers as shown in Figure~\ref{fig:model_maker_pipeline}.
This model maps the gesture embedding vectors generated by the pre-trained \textit{single-hand embedding} model to the target labels of the input images.

To train the custom gesture recognition model, users need to supply a small set of images.
Each image should be annotated with a hand gesture label.
All input images are pre-processed by the hand landmark model to generate hand landmarks on the fly during model training.

Our training pipeline allows the users to customize the neural network attributes such as the dense layer shapes and the training hyperparameters such as learning rate, batch size, and training epochs, etc.
Because of low training data requirements, each training session only takes a few minutes on most local computers and on Google's public Colab~\cite{colab} runtime to produce accurate gesture recognition models.
The trained custom gesture model is then converted to a TFLite \cite{tflite} model format for our end-to-end inference pipeline introduced below.

\subsection{Inference Pipeline}

The gesture recognition inference pipeline has been implemented as a modular structure, as shown in Figure~\ref{fig:model_maker_pipeline}. 

The pipeline consumes a raw hand image sequence as input and processes all images sequentially.
The hand landmark detection module converts the input images into landmark vectors.
The gesture embedding module further maps the landmark vectors to 128-dimensional gesture embedding vectors.
The gesture recognition module outputs the probabilities of each label.
This modular graph structure allows users to control or replace any module as desired.

Our benchmarks show that this end-to-end pipeline achieves real time performance ($16.76~ms$ per frame) on Pixel $6$ devices.

Our inference pipeline \textit{MediaPipe Tasks}~\cite{mediapipe_tasks}
offers a user-friendly API that supports multiple platforms, including Java, Python, and WebJs. This API allows users to easily integrate their customized gesture recognition model into the pipeline.

Both the training and inference pipeline have been open-sourced via \textit{MediaPipe Model Maker}~\cite{mediapipe_model_maker} and \textit{Gesture Recognizer API} in \textit{MediaPipe Tasks}~\cite{mediapipe_tasks}.

\section {Conclusion}
In conclusion, our research presents an easy-to-use approach to train accurate custom hand gesture recognition models with just a small set of training examples by fine-tuning pre-trained embeddings of hand landmarks. We also present our improvements to the hand landmark model which enhance the effectiveness of our hand gesture recognition system.  These findings underscore the practicality of our custom hand gesture recognition system in real world scenarios and pave the way for better human-computer interactions in various domains, such as virtual reality, augmented reality, video conferencing and remote control applications.

{\small
\nocite{ioffe2015batch}
\bibliographystyle{unsrt}
\bibliography{iccv2023_cv4metaverse_workshop_paper_accepted}
}

\end{document}